\documentclass[sigconf, screen, nonacm]{acmart}  

\usepackage{booktabs}       
\usepackage{array}          
\usepackage{caption}        
\usepackage{adjustbox}      
\usepackage{float}          

\usepackage{multirow}  
\usepackage{booktabs}  
\usepackage{makecell}  
\usepackage{geometry}  
\usepackage{placeins}

\usepackage{listings}                
\usepackage{xcolor}                  
\usepackage{enumitem}                
\usepackage{mathrsfs}
\usepackage[utf8]{inputenc}
\usepackage{graphicx}
\usepackage{stfloats}  
\newcommand{\std}[1]{\scriptsize$\pm$#1}

\usepackage{subcaption}  

\usepackage{algorithm}
\usepackage{algorithmic}
\usepackage{amsmath}
\usepackage{amsfonts}

\AtBeginDocument{%
  }

\setcopyright{acmlicensed}
\copyrightyear{2018}
\acmYear{2018}
\acmDOI{XXXXXXX.XXXXXXX}
\acmConference[Conference acronym 'XX]{Make sure to enter the correct
  conference title from your rights confirmation email}{June 03--05,
  2018}{Woodstock, NY}
\acmISBN{978-1-4503-XXXX-X/2018/06}

\begin{document}

\title{AgentLTV: An Agent-Based Unified Search-and-Evolution Framework for Automated Lifetime Value Prediction}


\author{Chaowei Wu}
\authornote{Both authors contributed equally to this research.} 
\orcid{0009-0007-4272-9165}
\affiliation{%
  \institution{Sichuan University}
  \city{Chengdu}
  \state{Sichuan}
  \country{China}
}
\email{wuchaowei@stu.scu.edu.cn}

\author{Huazhu Chen}
\orcid{0009-0007-4272-9165}
\authornotemark[1]
\orcid{}
\affiliation{%
  \institution{Harbin Institute of Technology}
  \city{Harbin}
  \state{Heilongjiang}
  \country{China}}
\email{chnghrr@gmail.com} 

\author{Congde Yuan}
\orcid{0000-0001-5298-3342}
\affiliation{%
  \institution{Sun Yat-sen University}
  \city{Guangzhou}
  \state{Guangdong}
  \country{China}}
\email{congdeyuan@gmail.com}

\author{Qirui Yang}
\orcid{0000-0003-0885-1830}
\affiliation{%
  \institution{City University of Hong Kong}
  \city{Hong Kong}
  \country{China}}
\email{evan.yang@my.cityu.edu.hk}

\author{Guoqing Song}
\orcid{0000-0002-8823-3540}
\affiliation{%
  \institution{VIVO}
  \city{Shenzhen}
  \state{Guangdong}
  \country{China}}
\email{karrysong666@gmail.com}

\author{Yue Gao}
\orcid{0009-0005-4460-0109}
\affiliation{%
\institution{Xiangtan University}
\city{Xiangtan}
\state{Hunan}
\country{China}
}
\email{202421050916@smail.xtu.edu.cn}

\author{Li Luo}
\orcid{0000-0002-2007-7916}
\affiliation{%
  \institution{Sichuan University}
  \city{Chengdu}
  \state{Sichuan}
  \country{China}}
\email{luolicc@scu.edu.cn}

\author{Frank Youhua Chen}
\orcid{0000-0003-4707-9361}
\affiliation{%
  \institution{City University of Hong Kong}
  \city{Hong Kong}
  \country{China}}
\email{youhchen@cityu.edu.hk}

\author{Mengzhuo Guo}
\authornote{Corresponding author.}
\orcid{}
\affiliation{%
  \institution{Sichuan University}
  \city{Chengdu}
  \state{Sichuan}
  \country{China}}
\email{mengzhguo@scu.edu.cn}

\renewcommand{\shortauthors}{Chaowei Wuet al.}


\begin{abstract}
Lifetime Value (LTV) prediction is critical in advertising, recommender systems, and e-commerce. In practice, LTV data patterns vary across decision scenarios. As a result, practitioners often build complex, scenario-specific pipelines and iterate over feature processing, objective design, and tuning. This process is expensive and hard to transfer. 
We propose \textbf{AgentLTV}, an agent-based unified search-and-evolution framework for automated LTV modeling. AgentLTV treats each candidate solution as an {executable pipeline program}. LLM-driven agents generate code, run and repair pipelines, and analyze execution feedback. Two decision agents coordinate a two-stage search. The Monte Carlo Tree Search (MCTS) stage explores a broad space of modeling choices under a fixed budget, guided by the Polynomial Upper Confidence bounds for Trees criterion and a Pareto-aware multi-metric value function. The Evolutionary Algorithm (EA) stage refines the best MCTS program via island-based evolution with crossover, mutation, and migration.
Experiments on a large-scale proprietary dataset and a public benchmark show that AgentLTV consistently discovers strong models across ranking and error metrics. Online bucket-level analysis further indicates improved ranking consistency and value calibration, especially for high-value and negative-LTV segments. We summarize practitioner-oriented takeaways: use MCTS for rapid adaptation to new data patterns, use EA for stable refinement, and validate deployment readiness with bucket-level ranking and calibration diagnostics. The proposed AgentLTV has been successfully deployed online.
\end{abstract}




\maketitle

\section{Introduction}
Lifetime Value (LTV) measures the net monetary value expected from a user or a customer \cite{borle2008customer}. Many studies have predicted one's LTV over a predefined period to guide subsequent business strategies, including online display bidding optimization \cite{su2024cross, chan2011measuring}, traffic allocation \cite{wang2024adsnet}, recommender systems \cite{pan2025progressive, weng2024optdist}, and long-term business planning \cite{liu2024mdan}. Accurate LTV predictions can help decision-makers target the high-value users and optimize profits. 

Current studies of LTV predictions can be classified into three streams. The first stream uses probabilistic models to capture users' purchase and churn behavior, assuming these behaviors follow specific distributions \cite{schmittlein1987counting, fader2005counting, jerath2011new, schmittlein1987counting}. However, such assumptions may simplify one's behaviors and thus reduce the model accuracy. The second stream is based on machine learning methods \cite{vanderveld2016engagement, win2020predicting}. They perform better than the first stream of methods, but cannot deal with large-scale data with high-dimensional features \cite{kim2025comprehensive}. The last stream utilizes deep learning methods with well-designed model structures to capture implicit and complex user behavioral patterns \cite{wang2024adsnet, li2022billion, su2024cross, chen2025mini}.


\begin{figure*}[!htbp]
  \centering
  \includegraphics[width=\linewidth]{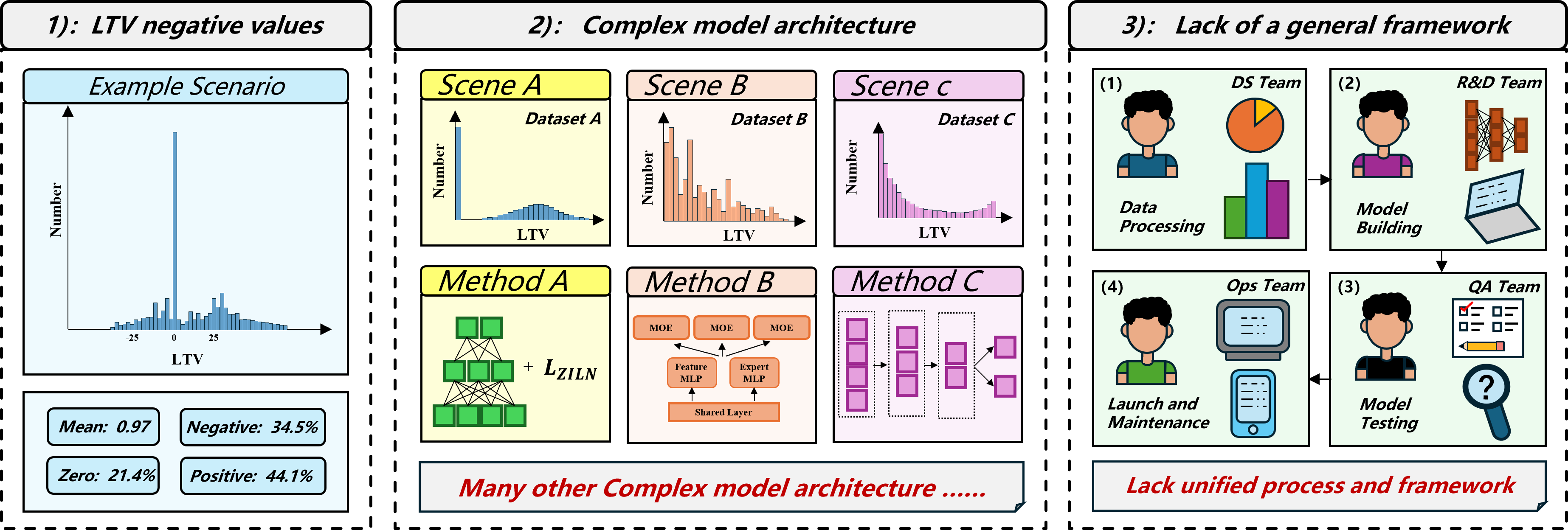}  
  \caption{The current challenges faced by the LTV prediction scene.}
  \Description{The current challenges}
\label{fig:Current_situation} 
\end{figure*}

Although deep learning has contributed to LTV prediction, these models remain hard to adopt in industry. We highlight three practical challenges as shown in Figure \ref{fig:Current_situation}. First, most existing methods do not explicitly handle negative LTV since they implement the popular zero-inflated log-normal (ZILN) loss \cite{wang2019deep, su2023cross, zhang2023out, xu2025hiltv}. LTV can be negative when the platform spends money on a user but receives little or no return. This happens, for example, in advertising when impressions incur cost but users do not click. It also arises in short-video platforms when cash-out incentives exceed user contributions. Negative values break common assumptions in LTV modeling. Second, building an LTV model still requires substantial manual effort.

Moreover, LTV prediction is often sparse, complex, and its data patterns are hard to capture. Both academic and industrial systems, therefore, tend to build highly customized and complex models, such as GNNs for social relations \cite{xing2021learning} or explicit multi-distribution mechanisms for heterogeneous payments \cite{weng2024optdist}. However, these designs usually come with a heavy pipeline. Practitioners must iterate over data cleaning, feature engineering, objective design, and hyper-parameter tuning. This leads to high development and maintenance costs. 

Finally, these complex models are rarely reusable across scenarios. They are typically optimized for a specific business setting or dataset. When a new scenario emerges, the previous model may overfit to old patterns and fail to transfer. Recent work uses LLMs to automate code generation \cite{zhang2025systematic}, but this does not fully solve the problem. If the LLM tries to mine deep context, it can inherit the same overfitting risk. If it only uses a wide context without LTV-specific signals, the generated pipeline is often misaligned with the task. In practice, code generation must balance deep, scenario-specific context and wide, generalizable context.

To address these challenges, we propose \textbf{AgentLTV}, an agent-based unified \emph{search-and-evolution} framework for automated LTV modeling. 
AgentLTV performs a two-stage automated search over {executable pipeline programs}. 
In the first stage, a Configuration module initializes multiple diverse root programs using dataset descriptions and LTV domain knowledge. An \textbf{Monte Carlo Tree Search (MCTS) Agent} then iteratively selects promising nodes using a Pareto-aware multi-metric value function together with the Polynomial Upper Confidence bounds for Trees (PUCT) criterion~\cite{silver2016mastering}. 
For each expansion, auxiliary agents generate candidate edits, synthesize runnable code, and execute/repair the pipeline to obtain reliable metrics. 
This stage outputs the best program $c_{\text{MCTS}}$ under a fixed budget.

In the second stage, an \textbf{Evolutionary Algorithm (EA) Agent} refines $c_{\text{MCTS}}$ via an island-based evolutionary process. 
AgentLTV deploys mutated variants of $c_{\text{MCTS}}$ to multiple islands, evolves them through selection and prompt-guided crossover/mutation, and periodically performs inter-island migration to share high-quality building blocks. 
The final output is a single deployable LTV pipeline program that is optimized for the target scenario.

We evaluate AgentLTV on large-scale proprietary industrial datasets and a public benchmark. 
Results show that AgentLTV consistently discovers strong LTV models while reducing engineering workload. 
We further conduct ablations and online economic analysis to quantify the contribution of each stage and to validate practical value in production.

Our contributions are summarized as follows:

\textbf{(1) Unified search-and-evolution framework for LTV modeling.} 
We introduce AgentLTV as a general framework that searches over LTV pipeline programs without assuming a fixed model family. 
By combining MCTS for broad exploration and EA for refinement, AgentLTV adapts to different data patterns and can be easily deployed to any LTV prediction scenario.

\textbf{(2) Execution-grounded multi-agent auto-coding for end-to-end automation.} 
AgentLTV is, to our knowledge, the first to apply an LLM-driven multi-agent system to LTV modeling. 
Two decision agents (MCTS/EA) orchestrate the process, while auxiliary agents generate code, run and repair pipelines, and propose grounded edits. 
This enables scalable automation in a large and structured decision space, reducing human intervention.

\textbf{(3) Practitioner-oriented guidance for LTV prediction.} 
Our proposed AgentLTV discovers deployable pipelines and handles negative LTV. 
It achieves the best offline metrics and improves online bucket-level ranking consistency and MAE, especially for high-value and negative-value users. 
Ablations show MCTS drives exploration and EA refines $c_{\text{MCTS}}$. 
Based on these results, we recommend practitioners using MCTS for exploring when data patterns or constraints change, and using EA for stable refinement under a fixed budget, while validating both ranking and calibration via bucket-level diagnostics before deployment. Our AgentLTV has been successfully deployed online.

\section{Related Work}
In this section, we briefly review some related work, mainly including Lifetime Value Prediction and Code Generation with LLMs.

\subsection{Lifetime Value Predictions}
Existing approaches to LTV prediction can be broadly categorized into three paradigms. The first line of work is based on probabilistic statistical models. 
Early representative methods include the RFM framework \cite{fader2005rfm}, which segments users by recency, frequency, and monetary value, and the Pareto/NBD model \cite{schmittlein1987counting}, which characterizes purchase arrivals via a Poisson process and customer churn via an exponential distribution. The Pareto/NBD model and its extensions \cite{fader2005counting, jerath2011new} form the core of the Buy-Till-You-Die (BTYD) family \cite{wadsworth2012buy}. While these models offer strong interpretability, they are limited in modeling long-term dynamics and fine-grained monetary outcomes.

The second category consists of machine learning–based methods, which typically frame LTV prediction as a regression or classification problem. Prior studies have applied tree-based models such as random forests for LTV prediction in retail and gaming scenarios \cite{win2020predicting, sifa2015predicting}. Vanderveld et al. \cite{vanderveld2016engagement} proposed a two-stage framework that separately models purchase propensity and spending amount. Although effective in structured settings, these methods often struggle with sequential dependencies and complex nonlinear behavioral patterns. 

The third category is based on deep learning methods. More recent advances are dominated by deep learning–based approaches, which explicitly model temporal dynamics and heterogeneous user behaviors. Chen et al. \cite{chen2018customer} demonstrated that CNN-based architectures are effective for identifying high-value “whale” users in games. To address the zero-inflated and heavy-tailed nature of LTV distributions, Wang et al. introduced the Zero-Inflated Log-Normal (ZILN) loss \cite{wang2019deep}, with subsequent work further enhancing temporal and structural modeling. Representative extensions incorporate wavelet-based temporal decomposition, recurrent networks, attention mechanisms, and graph neural networks to learn robust sequential and relational representations \cite{xing2021learning, xu2025hiltv}.In industrial-scale applications, LTV modeling has evolved toward production-oriented, multi-task, and distribution-aware systems. Several large platforms have reported deploying LTV solutions, including Kuaishou’s billion-user LTV system \cite{li2022billion}, NetEase’s perCLTV framework \cite{zhao2023percltv}, OPPO’s progressive task cascading model PTMSN \cite{pan2025progressive}, and Baidu’s uncertainty-aware multi-task learning approach \cite{yang2025use}. Tencent further proposed influential systems such as ExpLTV \cite{zhang2023out} and adaptive multi-source feature fusion for improving long-tail performance \cite{yang2023feature}. More recently, OptDist \cite{weng2024optdist} and staged multi-task optimization frameworks deployed in WeChat \cite{chen2025mini} have demonstrated strong practical effectiveness.

Despite substantial progress, most existing methods are designed to address isolated challenges such as data sparsity, distribution skewness, or feature interactions. In practice, this often requires significant manual effort to engineer complex, scenario-specific models, resulting in long development cycles and limited transferability. Consequently, a general and automated LTV modeling framework that can adapt across datasets and application scenarios remains an open and important research direction.

\subsection{Code Generation with LLMs}
Recent advances in large language models have led to two dominant paradigms for code generation. The first paradigm follows a breadth-oriented strategy, where LLMs are prompted to produce a large number of independent candidate programs, and the best solution is selected via execution-based or heuristic evaluation. 
Representative works in this line include \cite{chen2021evaluating, austin2021program, hendrycks2021measuring}. However, in complex scenarios such as LTV modeling, simple prompt-based generation often fails to produce executable and effective programs. To address such a limitation, subsequent studies leverage execution feedback to iteratively refine and repair generated code, enabling self-correction and progressive quality improvement \cite{zhang2023self, welleck2022generating, madaan2023self}. Building on this idea, tree search–based methods have been introduced to guide code generation over structured solution spaces. Yao et al. \cite{yao2023tree} and Koh et al. \cite{koh2024tree} combined LLMs with tree search to support complex reasoning in NLP and multimodal tasks, while Google demonstrated the effectiveness of this paradigm for expert-level programming problems at scale \cite{aygun2025ai}. A complementary paradigm focuses on depth-oriented code evolution, where LLMs guide the refinement and mutation of programs over multiple iterations. FunSearch \cite{romera2024mathematical} showed that LLM-guided evolutionary search is effective for discovering mathematical functions and combinatorial structures, and AlphaEvolve \cite{novikov2025alphaevolve} extended this approach to general program optimization and algorithm design. More recent systems, such as AlphaResearch \cite{yu2025alpharesearch}, further integrate execution-based verification with simulated peer review signals, while CodeEvolve \cite{assumpccao2025codeevolve} and ShinkaEvolve \cite{lange2025shinkaevolve} improve scalability through evolutionary strategies and adaptive LLM scheduling.

Despite these advances, directly applying existing LLM-based code generation paradigms to LTV modeling remains challenging. Breadth-oriented methods often replicate existing open-source pipelines without producing runnable and accurate predictive models, while depth-oriented evolutionary approaches typically assume the availability of a stable initial program, which is difficult to obtain in realistic LTV settings. Consequently, effectively balancing broad exploration and structured refinement remains a key challenge for automatic LTV model code generation.

\section{Problem Definition}
We study LTV prediction, which aims to estimate a user’s long-term value from multi-dimensional features. Formally, we are given a labeled dataset
\[
\mathcal{D} = \left\{ (x_i, y_i) \right\}_{i=1}^N,
\]
where $x_i\in\mathcal{X}$ is the feature vector of user $i$, and $y_i\in\mathbb{R}$ is the ground-truth LTV label (which may be negative). The goal is to learn a prediction function $f$ parameterized by $\Theta$ that maps features to the predicted LTV:
\begin{equation}
\widehat{y}_i=\mathrm{PLTV}_i=f(x_i;\Theta).
\end{equation}
We use $\mathcal{F}$ to denote the space of candidate modeling pipelines (i.e., choices of feature transformations, model components, and training recipes) that define $f$. Rather than manually specifying $f$, we introduce two automated modules to search for a high-performing pipeline in $\mathcal{F}$. Details are provided in the next section.


\section{Framework of AgentLTV}

\begin{figure*}[!htbp]  
  \centering
  \includegraphics[width=\linewidth]{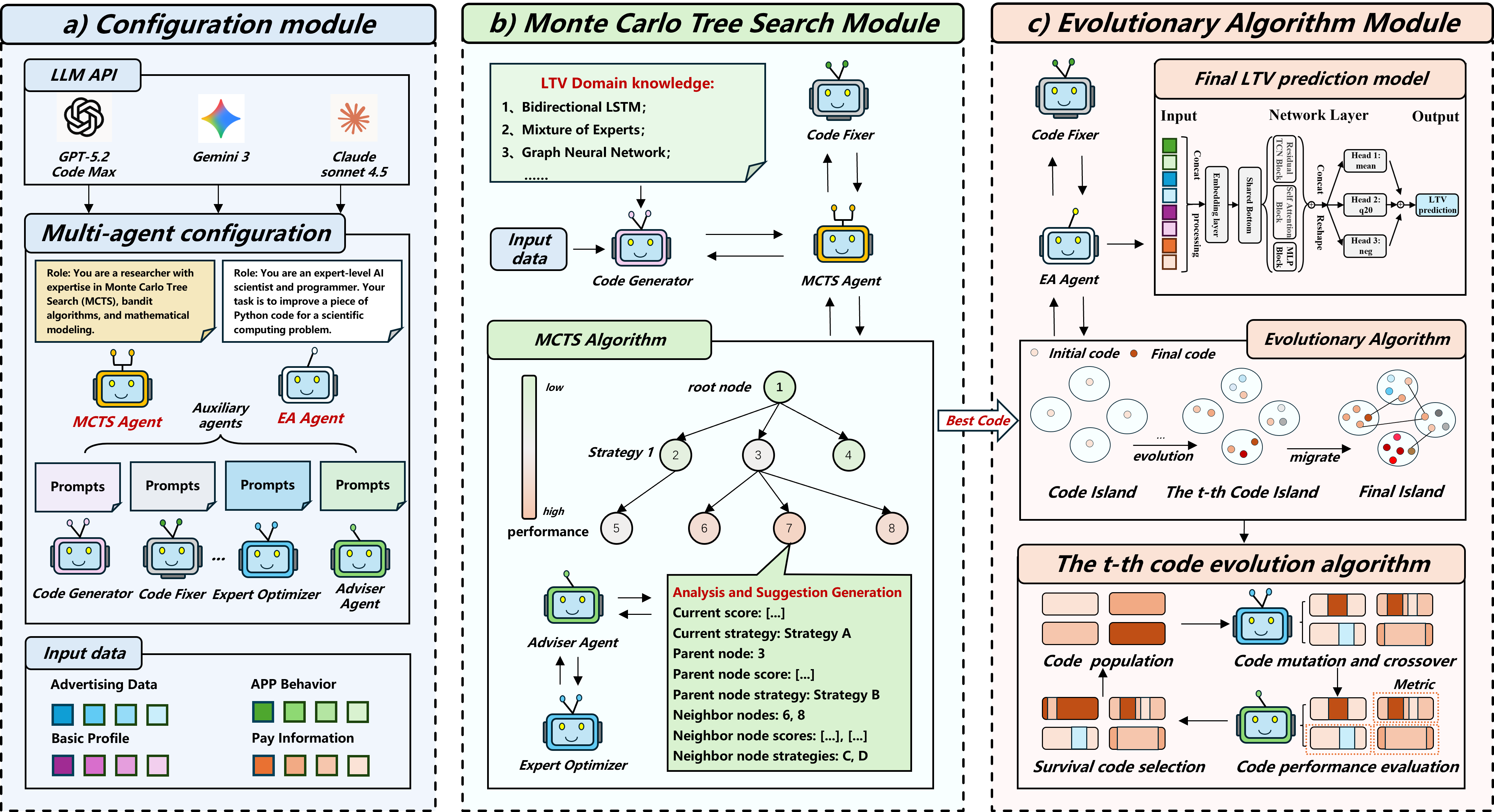}
  \caption{AgentLTV overview. (a) Configuration module instantiates LLM-driven agents and execution interfaces. 
(b) MCTS module performs broad exploration over LTV pipeline programs via PUCT-guided expansion after the Code Generator generates the initial node algorithm. 
(c) EA module refines the best MCTS program using island-based evolution with crossover, mutation, and migration based on the previous optimal algorithm.}
  \Description{Model Architecture}
\label{fig:model_arch} 
\end{figure*}

As shown in Figure~\ref{fig:model_arch}, we propose \textbf{AgentLTV}, an automated LTV modeling framework based on LLM-driven multi-agent collaboration. AgentLTV has three tightly coupled modules. The Configuration module instantiates decision agents (e.g., MCTS and EA agents) and auxiliary agents (e.g., Code Generator, Code Fixer, Expert Optimizer, and Adviser Agent). The MCTS module performs broad exploration over high-level modeling decisions. The EA module further refines the best program from MCTS via evolutionary search and outputs the optimal LTV prediction code. We next introduce these modules in detail.

\subsection{Configuration Module}
The Configuration module instantiates a team of LLM-driven agents using role-specific prompts and different LLM APIs (e.g., GPT,\footnote {https://openai.com/} Claude,\footnote {https://claude.ai} Gemini\footnote {https://gemini3-pro.com/}).
The decision agents are the \textbf{MCTS Agent} and the \textbf{EA Agent}. They decide which candidate program to expand (MCTS) and how to refine and select programs via evolution (EA). All other agents are auxiliary agents. All agents follow a unified pipeline interface (data loading, training, evaluation, and logging), which enables fair comparison between candidate programs.
The prompts are provided in Appendix~\ref{app:Agent Prompts}.

\begin{itemize}
    \item \textbf{Code Generator.} Given structured prompts (edits, constraints, and interfaces), it generates or modifies executable pipeline code (i.e, program).
    \item \textbf{Code Fixer.} It serves for instantiating candidate codes, managing training and evaluation, handling execution failures, and performing iterative repair to ensure each candidate program is executable and comparable.
    \item \textbf{Adviser Agent.}
It analyzes program structure and execution results and autonomously proposes optimization suggestions, including static human knowledge, reorganization of high-performance solutions, and innovative research methods based on fundamental principles.
\item \textbf{Expert Optimizer.}
It injects LTV domain knowledge (e.g., common architectures and loss/target transforms) by selecting the most relevant guidance. It then converts suggestions into actionable, interface-consistent prompts to generate new solution code.
\item \textbf{MCTS Agent.}
It implements the MCTS loop and selects which node (i.e., program) to expand using the PUCT criterion. It use the Code Generator to propose a child program and the Code Fixer to execute and repair it. It also uses priors and suggestions produced by the Adviser Agent and Expert Optimizer.
\item  \textbf{EA Agent.}
It implements island-based evolutionary search. It invokes multiple agents to ensure the evolution, evaluation, selection, and cross-island migration of code to produce optimal code.
\end{itemize}

\begin{algorithm}[!htbp]
\caption{The Algorithm of MCTS Module} 
\label{alg:mcts_ltv}
\fontsize{8pt}{10pt}\selectfont 
\begin{algorithmic}[1]
\REQUIRE Dataset $\mathcal{D}$, $c_{\mathrm{puct}}$, roots $K$, search budget $B$
\ENSURE Search LTV prediction program $c_{\mathrm{MCTS}}$

\STATE $\mathcal{T} \leftarrow \emptyset$\\
\COMMENT{(1) Multi-Root Initialization}
\FOR{$k = 1$ to $K$} 
    \STATE Generate $c_{k}^{(0)}$ via \textbf{Code Generator}
    \STATE $\mathbf{o}(c_{k}^{(0)}) \leftarrow$ Evaluate $c_{k}^{(0)}$ on $\mathcal{D}$ via \textbf{Code Fixer}
    \STATE $R(c_{k}^{(0)}) \leftarrow \sum_i w_i \cdot \mathrm{norm}(o_i(c_{k}^{(0)})) + \beta\cdot\mathbb{I}_{\text{Pareto}}(c_{k}^{(0)}) + \gamma \cdot d_{\text{crowd}}(c_{k}^{(0)})$
    \STATE Initialize $N(c_{k}^{(0)})\leftarrow 1$, $P(c_{k}^{(0)})$, $Q(c_{k}^{(0)}) \leftarrow R(c_{k}^{(0)})$
    \STATE Add $c_{k}^{(0)}$ to $\mathcal{T}$ as a root
\ENDFOR\\
\COMMENT{(2) Node selection with PUCT}
\FOR{$b = 1$ to $B$}
    \STATE $c^* \leftarrow \arg\max_{c \in \mathcal{T}} \left( Q(c) + c_{\mathrm{puct}} \cdot P(c) \frac{\sqrt{  N(u)}}{1 + N(c)} \right)$\\
    \COMMENT{(3) LLM-driven expansion}
    \STATE Generate suggestions for expanding $c^*$ via \textbf{Advice Generator}
    \STATE Calculate $P(\cdot)$ for candidate suggestions via \textbf{Expert Optimizer}
    \STATE Generate a child program $c_{\text{new}}$ via \textbf{Code Generator}
    \STATE $\mathbf{o}(c_{\text{new}}) \leftarrow$ Evaluate $c_{\text{new}}$ on $\mathcal{D}$ via \textbf{Code Fixer}
    \STATE $Q(c_{\text{new}}) \leftarrow \sum_i w_i \cdot \mathrm{norm}(o_i(c_{\text{new}})) + \beta\cdot\mathbb{I}_{\text{Pareto}}(c_{\text{new}}) + \gamma \cdot d_{\text{crowd}}(c_{\text{new}})$
    \STATE $N(c_{\text{new}}) \leftarrow 1$
    \STATE Calculate $P(c_{\text{new}})$ as the weighted assessment
    \STATE Add $c_{\text{new}}$ to $\mathcal{T}$ as a child of $c^*$

    \COMMENT{(4) Backpropagation and output}
    \FOR{all $c_a \in \mathrm{ancestor}(c_{\text{new}})$}
    \STATE $N(c_{\text{new}}) \leftarrow N(c_{\text{new}}) + 1$
    \STATE $Q(c_{\text{new}}) \leftarrow \frac{Q(c_{\text{new}})\cdot N(c_{\text{new}}) + R(c_a)}{N(c_{\text{new}})}$
\ENDFOR
\ENDFOR\\
\COMMENT{(4) Backpropagation and output}
\STATE $c_{\mathrm{MCTS}} \leftarrow \arg\max_{c \in \mathcal{T}} Q(c)$ 
\RETURN $c_{\mathrm{MCTS}}$
\end{algorithmic}
\fontsize{10pt}{12pt}\selectfont  
\end{algorithm}

\subsection{Monte Carlo Tree Search Module}

This module aims to identify the optimal LTV prediction code using the MCTS algorithm, with multiple agents initialized by the Configuration module. Algorithm~\ref{alg:mcts_ltv} summarizes the MCTS module.

The MCTS module searches for a high-performing LTV pipeline program under a fixed execution budget.
We denote a complete pipeline by a program $c$, which instantiates a predictor $f$ together with feature processing and training recipes. The search space is $\mathcal{F}$, and explored programs form a tree $\mathcal{T}$. The workflow of MCTS module is as follows:

\textbf{(1) Multi-root initialization.}
We initialize $K$ diverse root programs $\{c_k^{(0)}\}_{k=1}^{K}$ by using Code Generator given method-specific prompts (see Appendix~\ref{app:Agent Prompts}).
Each root $c_{k}^{(0)}$ represents an LTV prediction model. It is executed and evaluated by the Code Fixer to obtain metrics.

\textbf{(2) Node selection with PUCT.}
Using the {Code Fixer} and the {MCTS Agent}, we execute each node program, collect its output, and compute the corresponding metrics. We then combine the \textbf{PUCT} criterion with a \textbf{Pareto-aware multi-objective reward} to balance exploration and exploitation. The node selection rule is:
\begin{equation}
c^* = \arg\max_{c \in \mathcal{T}} \left( Q(c) + c_{\mathrm{puct}} \cdot P(c) \frac{\sqrt{N(u)}}{1 + N(c)} \right),
\end{equation}
where $Q(c)$ is the value of node $c$, $V(c)$ (resp. $V(u)$) is the node (resp. parent nodes of $c$) visit count. $c_{\mathrm{puct}}$ is the exploration coefficient. For each candidate expansion (i.e., child node), the {Advice Generator} produces a set of optimization suggestions by analyzing the current node and its context (e.g., parent and neighboring nodes) along with execution metrics. The {Expert Optimizer} then scores and reweights these suggestions based on LTV domain knowledge and interface constraints (We present this LTV domain knowledge in Appendix~\ref{app:LTV Domain knowledge}). We denote this weighted assessment as $P(c)$, which serves as a prior to guide which candidate node to explore. Given the selected suggestion, the {Code Generator} produces the corresponding child-node program $c$.

Considering that standard MCTS typically updates node values using a single scalar metric, while LTV prediction is usually evaluated by multiple criteria, such as ranking- and accuracy-related metrics, we update node values using a Pareto-aware reward function:
\begin{equation}
R(c) = \sum_i w_i \cdot \mathrm{norm}(o_i(c)) + \beta \cdot \mathbb{I}_{\text{Pareto}}(c) + \gamma \cdot d_{\text{crowd}}(c),
\end{equation}
where $o_i(c)$ denotes the $i$-th evaluation metric of node $c$, $\mathrm{norm}(\cdot)$ is a normalization operator, $\mathbb{I}_{\text{Pareto}}(c)$ indicates whether $c$ lies on the Pareto frontier, $d_{\text{crowd}}(c)$ is a diversity reward based on crowding distance, and $w_i$, $\beta$, and $\gamma$ are coefficients. Next, node values are updated by incremental averaging.


\textbf{(3) LLM-driven expansion.}
Given $c^*$, the Adviser Agent analyzes $c^*$ as well as its parent and neighbors, and proposes candidate edits. The Expert Optimizer grounds these edits with LTV knowledge and converts them into interface-consistent prompts. The Code Generator produces a new child program, which the Code Fixer executes (and repairs if needed).


\textbf{(4) Backpropagation and output.}
The evaluation results are backpropagated to update node statistics.
After exhausting the search budget, we return the best program.
\begin{equation}
c_{\mathrm{MCTS}}=\arg\max_{c\in\mathcal{T}}Q(c).\end{equation}


\subsection{Evolutionary Algorithm Module}
The EA module further refines the best program produced by the MCTS module and outputs the final LTV prediction code. 
Algorithm~\ref{alg:code_evolution} summarizes the procedure.

\textbf{(1) Multi-island initialization.}
We maintain $K$ independent evolutionary islands to promote diversity and reduce premature convergence:
\begin{equation}
\mathcal{I}=\{P_1,P_2,\ldots,P_K\}.
\end{equation}
We use the optimal program  $c_{\text{MCTS}}$ as the initial population code. EA agent uses Code Generator to create $K$ mutated variants and uses them to initialize island populations $\{P_k^{(0)}\}_{k=1}^{K}$.


\textbf{(2) Code evolution.}
Each island evolves for $T$ generations. 
At generation $t$, we first evaluate every program in $P_k^{(t-1)}$ on $\mathcal{D}$ to obtain its fitness $F(c)$, while discarding non-executable programs via the Code Fixer. 
We then select an elite subset $E_k^{(t)}$ according to the elite ratio $\rho$ and the fitness ranking.
Next, we sample parent programs from $E_k^{(t)}$ and generate offspring via crossover and mutation.
The Advice Generator and Expert Optimizer provide structured prompts for these operations. 
Mutations cover feature augmentation, architecture refinement, hyperparameter tuning, and loss substitution. 
Finally, we form the next generation, $P_k^{(t)}$, by combining elites with newly generated individuals and removing other programs to indicate ``population death''.


\textbf{(3) Inter-island migration and global selection.}
Periodic migration every $\tau$ generations transfers high-fitness programs between islands to enable knowledge sharing while preserving diversity.
After $T$ generations, we select the globally best program from all islands:
\begin{equation}
c^*=\arg\max_{c\in \cup_{k=1}^{K} P_k^{(T)}} F(c).
\end{equation}


\begin{algorithm}[H]
\caption{The Algorithm of EA Module}
\label{alg:code_evolution}
\fontsize{8pt}{10pt}\selectfont 
\begin{algorithmic}[1]  
\REQUIRE Program $c_{\text{MCTS}}$, dataset $\mathcal{D}$, number of islands $K$, generations $T$, elite ratio $\rho$, migration period $\tau$
\ENSURE Globally optimal LTV prediction program $c^*$

\STATE $\mathcal{I} \leftarrow \emptyset$

\COMMENT{(1) Multi-island initialization (seeded by $c_{\text{MCTS}}$)}
\FOR{$k \leftarrow 1$ \TO $K$}
    \STATE $P_k^{(0)} \leftarrow \text{Mutate}(c_{\text{MCTS}}; \mathcal{D})$ \COMMENT{via EA Agent with \textbf{Adviser Agent}, \textbf{Expert Optimizer}, and \textbf{Code Generator}}
    \STATE $\mathcal{I} \leftarrow \mathcal{I} \cup \{P_k^{(0)}\}$
\ENDFOR
\COMMENT{(2) Code evolution}
\FOR{$t \leftarrow 1$ \TO $T$}
    \FOR{all islands $P_k^{(t-1)} \in \mathcal{I}$}
        \FOR{each $c \in P_k^{(t-1)}$}
            \STATE $F(c) \leftarrow \text{Eval}(c; \mathcal{D})$ \COMMENT{via \textbf{Code Fixer}}
        \ENDFOR
        \STATE $P_k^{(t-1)} \leftarrow \{c \in P_k^{(t-1)} \mid c \text{ executable}\}$ \COMMENT{examined by \textbf{Code Fixer}}

        \STATE $E_k^{(t)} \leftarrow \text{SelectTop}(P_k^{(t-1)}, \rho)$

        \STATE $\mathcal{S} \leftarrow \text{Adviser}(E_k^{(t)})$ \COMMENT{Adviser Generator analyzes code + metrics}
        \STATE $\mathcal{S} \leftarrow \text{ExpertOptimize}(\mathcal{S})$ \COMMENT{Expert Optimizer Suggestions}
        \STATE $C_{\text{child}} \leftarrow \text{Crossover}(E_k^{(t)}; \mathcal{S})$ \COMMENT{EA Agent selects parents; \textbf{Code Generator} produces children}
        \STATE $O_k^{(t)} \leftarrow \{\text{Mutate}(c; \mathcal{S}) \mid c \in C_{\text{child}}\}$ \COMMENT{\textbf{Code Generator} edits}

        \STATE $P_k^{(t)} \leftarrow O_k^{(t)} \cup E_k^{(t)}$
    \ENDFOR

    \COMMENT{(3) Inter-island migration and Global selection}
    \IF{$t \bmod \tau = 0$}
        \FOR{all islands $P_k^{(t)} \in \mathcal{I}$}
            \STATE $P_k^{(t)} \leftarrow P_k^{(t)} \cup \{c \mid c \in E_{k'}^{(t)}, k' \neq k\}$
        \ENDFOR
    \ENDIF
\ENDFOR
\STATE $c^* \leftarrow \arg\max_{c \in \cup_{k=1}^{K} P_k^{(T)}} F(c)$
\RETURN $c^*$
\end{algorithmic}  
\fontsize{10pt}{12pt}\selectfont  
\end{algorithm}




\section{Experiment}
We conducted the following four experiments to evaluate AgentLTV. Unlike standard LTV predictors, AgentLTV is an automated framework that searches for executable LTV pipelines using MCTS exploration and EA refinement. 
We answer four research questions:
(RQ1) How does AgentLTV compare with LTV baselines? (RQ2) How do the MCTS and EA modules contribute to performance? (RQ3) How does the discovered model perform online? (RQ4) How well does AgentLTV generalize to a public benchmark?

\subsection{Experimental Setup}
\subsubsection{Dataset}
We run offline experiments on a large-scale industrial dataset from a popular mobile game (company details are omitted due to policy). 
The task is to predict the cumulative revenue within the next 60 days using the first 7 days after installation. 
We use data from two quarters, covering about 800{,}000 users. Some statistics of the data are shown in Table \ref{tab:data statistics}.
The dataset is split into train/validation/test with an 8:1:1 ratio for each period.

\begin{table}[!htbp]
  \caption{Basic statistics of each period dataset.}
  \label{tab:data statistics}{\footnotesize
  \begin{tabular}{ccccc}
    \toprule
    Period& Samples& Positive & Negative& Average LTV\\
    \midrule
    Period 1& 426,412 & 188,047& 146,686& 23.810\\
    Period 2& 382,081 & 171,554& 132,201& 25.323\\
  \bottomrule
\end{tabular}}
\end{table}

The features include three groups: user attributes, advertising attributes, and game behaviors. 
User features include age, gender, region, and city. 
Advertising features include channel, ad type, and install information. 
Game features include activity signals, competition scores/ranks, and behavior in different match modes. 
We use 132 features in total. 
The label $y$ is the realized 60-day revenue after installation, which can be negative due to cost-dominated users.


\subsubsection{Metrics}
We report four metrics: Error Rate (ER), Normalized Gini (Norm GINI), Spearman Rank Correlation (Spearman), and RMSE. 
We define
$\text{ER} = \frac{\sum_{i=1}^{n} \left| \hat{y}_i - y_i \right|}{\sum_{i=1}^{n} |y_i|}$,
$\text{Norm GINI} = \frac{2 \sum_{i=1}^{n} (r_{\hat{y},i} \cdot r_{y,i}) - n(n+1)}{n(n-1)}$,
$\text{Spearman} = 1 - \frac{6 \sum_{i=1}^{n} (r_{\hat{y},i} - r_{y,i})^2}{n(n^2 - 1)}$,
and $\text{RMSE} = \sqrt{\frac{1}{n} \sum_{i=1}^{n} (\hat{y}_i - y_i)^2}$,
where $r_{\hat{y},i}$ and $r_{y,i}$ are the ranks of $\hat{y}_i$ and $y_i$.


\subsubsection{Baselines}
We compare our model with the following representative baselines:
\textbf{Wide\&Deep}~\cite{cheng2016wide}: A classic industrial model that jointly captures memorization and generalization via parallel linear and deep components.
\textbf{DeepFM}~\cite{guo2017deepfm}: Combines factorization machines and DNNs to model both low- and high-order feature interactions in sparse settings.
\textbf{ZILN}~\cite{wang2019deep}: Addresses zero-inflated and long-tailed LTV distributions through zero-inflated log-normal modeling.
\textbf{DCN}~\cite{wang2017deep}: Explicitly models feature crosses with a Cross Network while capturing higher-order interactions via a Deep Network.
\textbf{GateNet}~\cite{huang2020gatenet}: Introduces gating mechanisms to adaptively filter informative features in high-dimensional sparse data.
\textbf{Kuaishou}~\cite{li2022billion}: Models highly imbalanced LTV distributions via value interval segmentation and multi-subdistribution learning.
\textbf{TSUR}~\cite{xing2021learning}: Use multi-channel discrete wavelet transform (DWT) with GRU and attention machine and graph attention network (GAT) predict LTV.
\textbf{OptDist}~\cite{weng2024optdist}: Learns adaptive sample-level distributions using multiple candidate subdistributions and Gumbel-Softmax selection.
\textbf{USE-LTV}~\cite{yang2025use}: Integrates uncertain behavior modeling with Transformer encoders and multi-gate expert architectures for long-tailed LTV prediction.
\textbf{HiLTV}~\cite{xu2025hiltv}: Proposes hierarchical distribution modeling with ZIMoL loss to capture zero, unimodal, and multimodal consumption patterns.

\subsubsection{Implementation Details}
We provide AgentLTV with an initial library of mainstream LTV modeling schemes as domain knowledge (Appendix~\ref{app:LTV Domain knowledge}). 
The MCTS module uses this knowledge to generate diverse root programs and then expands the search tree using execution feedback. 
We stop MCTS after obtaining 100 feasible and executable node programs.
Next, we initialize the EA module with $c_{\text{MCTS}}$, use 4 islands, and stop EA after generating 100 additional feasible programs.

AgentLTV have used multiple LLM APIs (claude-sonnet-4.5, gpt-5.2, and gemini-3-pro) for code generation and refinement. 
All experiments are conducted on a Linux server with two NVIDIA Tesla A100 GPUs and 128GB RAM.


\subsection{Optimal Model Explanation}

AgentLTV explores 133 nodes in the MCTS stage and evaluates 116 additional candidates in the EA stage. Figure~\ref{fig:tree_structure} visualizes the MCTS results, and table~\ref{tab:bt_ios_pltv_top20} in Appendix~\ref{app:search result} lists the top-20 candidate programs from EA stages. 
Algorithm~\ref{alg:AgentLTV} in Appendix~\ref{app:our best model} provide a detailed description of the final discovered model.

\begin{figure}[!htbp]
  \centering
  \includegraphics[width=\linewidth]{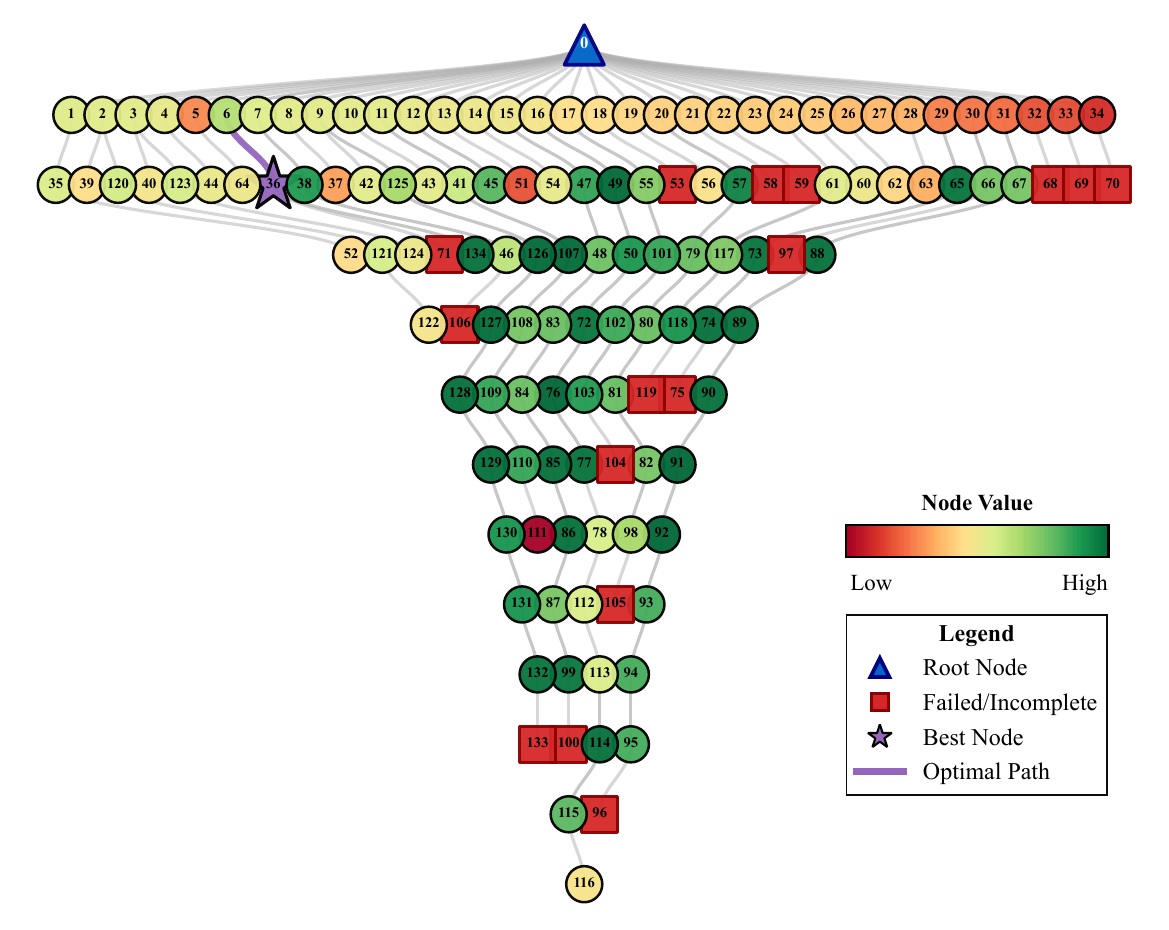}
  \caption{Monte Carlo tree search Module visualization.}
  \Description{Visualization of MCTS module. Note that node indices in the first layer reflect the prior ordering of initial programs: smaller indices correspond to more mainstream deep learning architectures, whereas larger indices represent earlier machine learning baselines.
}
  \label{fig:tree_structure}
\end{figure}

Figure~\ref{fig:tree_structure} shows a wide-but-shallow search pattern. This is expected in our setting because the first-layer nodes are domain-informed initial programs, and each expansion step applies a small set of program transformations. Under a limited evaluation budget and a large branching factor, the MCTS policy prioritizes breadth to quickly identify promising model families, rather than repeatedly deepening a single branch. 

Interestingly, the best-performing program emerges at depth 3. This indicates that, for LTV modeling, most gains come from a few high-impact structural decisions (e.g., model family and key design choices), while deeper expansions tend to correspond to finer-grained modifications with diminishing and noisier returns. Therefore, we treat MCTS primarily as a structure-level navigator and rely on the subsequent EA stage for further candidate filtering and refinement.

\subsection{Comparative Experiment}

\begin{table*}[!htbp]
  \caption{Performance comparison with different approaches}
  \label{tab: Performance}
  \begin{tabular}{ccccccccc}
    \toprule
     \multirow{2}{*}{\makecell{Model}} & \multicolumn{4}{c}{Period 1} & \multicolumn{4}{c}{Period 2} \\
    \cmidrule{2-9}
      & ER & Norm GINI & Spearman & RMSE & ER & Norm GINI & Spearman & RMSE \\
    \midrule
     Wide\&Deep 
      & 2.225\std{0.097} & 0.623\std{0.019} & 0.101\std{0.036} & 264.92\std{2.43}
      & 1.976\std{0.059} & 0.648\std{0.029} & 0.093\std{0.029} & 274.01\std{1.75} \\
     DeepFM 
      & 1.602\std{0.098} & 0.409\std{0.162} & 0.186\std{0.072} & 264.48\std{3.15}
      & 1.731\std{0.222} & 0.384\std{0.242} & 0.133\std{0.104} & 291.07\std{7.77} \\
     ZILN 
      & 1.492\std{0.174} & 0.310\std{0.450} & 0.323\std{0.216} & 261.83\std{4.07}
      & 1.719\std{0.161} & 0.034\std{0.311} & 0.189\std{0.161} & 287.64\std{1.69} \\
     DCN 
      & 2.625\std{0.166} & 0.602\std{0.025} & 0.030\std{0.049} & 284.11\std{9.87}
      & 2.392\std{0.126} & 0.606\std{0.037} & 0.028\std{0.036} & 295.41\std{9.84} \\
     GateNet 
      & 1.949\std{0.116} & 0.661\std{0.018} & 0.147\std{0.020} & 257.24\std{3.68}
      & 1.804\std{0.044} & 0.689\std{0.015} & 0.201\std{0.013} & 270.69\std{3.57} \\
     Kuaishou 
      & 1.894\std{0.140} & 0.681\std{0.084} & 0.523\std{0.063} & 261.69\std{2.49}
      & 2.123\std{0.068} & 0.580\std{0.086} & 0.493\std{0.036} & 287.83\std{1.57} \\
     TSUR 
      & 0.922\std{0.092} & 0.670\std{0.266} & 0.680\std{0.003} & 223.95\std{14.78}
      & 1.173\std{0.001} & 0.804\std{0.002} & 0.668\std{0.001} & 288.19\std{0.22} \\
     OptDist 
      & 1.418\std{0.059} & 0.775\std{0.022} & 0.515\std{0.044} & 258.03\std{8.93}
      & 1.526\std{0.089} & 0.770\std{0.053} & 0.454\std{0.032} & 284.63\std{0.48} \\
     USE-LTV 
      & 1.461\std{0.048} & 0.732\std{0.028} & 0.483\std{0.034} & 261.27\std{0.38}
      & 1.559\std{0.072} & 0.736\std{0.029} & 0.458\std{0.028} & 284.58\std{0.31} \\
     Hi-LTV 
      & 1.355\std{0.067} & 0.813\std{0.009} & 0.574\std{0.029} & 256.69\std{0.48}
      & 1.337\std{0.041} & 0.853\std{0.009} & 0.603\std{0.017} & 278.32\std{0.29} \\
     \textbf{AgentLTV}
      & \textbf{0.705\std{0.162}} & \textbf{0.963\std{0.008}} & \textbf{0.832\std{0.014}} & \textbf{133.58\std{28.23}}
      & \textbf{0.875\std{0.138}} & \textbf{0.959\std{0.005}} & \textbf{0.826\std{0.017}} & \textbf{145.61\std{16.24}} \\
    \bottomrule
  \end{tabular}
\end{table*}

To answer RQ1, we compare AgentLTV with representative LTV baselines. 
Table~\ref{tab: Performance} reports the results on two periods. 
AgentLTV achieves the best performance across all four metrics in both settings. 
Compared with the strongest baseline, AgentLTV reduces ER by 41.26\% on average, improves Norm GINI by 15.44\%, improves Spearman by 40.97\%, and reduces RMSE by 47.82\%.

These gains come from the automated search procedure. 
MCTS explores a broad space of modeling pipelines, and EA further refines promising candidates. 
The Pareto-aware reward helps balance ranking-oriented metrics and error-oriented metrics during search. 
In our dataset, the best discovered model includes temporal convolution and attention branches for dynamic behaviors, together with negative-aware output design and calibration, which improves both ranking quality and prediction accuracy.


\subsection{Ablation Study}

To answer RQ2, we ablate the two main search modules. 
Table~\ref{tab:Ablation} reports the results. 
Removing the EA stage changes performance noticeably, which shows that evolutionary refinement can further improve the MCTS-discovered solution in our setting. 
Removing the MCTS stage leads to a much larger drop in ranking-related metrics, indicating that broad exploration is critical for finding scenario-specific LTV pipelines in a large search space.

\begin{table}[H]
  \caption{The ablation study of MCTS Module and EA Module.}
  \label{tab:Ablation}{\footnotesize
  \begin{tabular}{ccccc}
    \toprule
    Method& ER& Norm GINI& Spearman& RMSE\\
    \midrule
    \begin{math} w/o  \end{math} MCTS& 1.030& 0.778& 0.633& 273.42\\
    \begin{math} w/o  \end{math} EA& 0.670& 0.958& 0.823& 185.95\\
    AgentLTV& 0.594& 0.962& 0.817& 141.09\\
  \bottomrule
\end{tabular}}
\end{table}

\subsection{Economic Analysis}

To evaluate online impact (RQ3), we conduct an analysis on real traffic logs from July 2025. 
We report (i) \textbf{Same-interval ratio}, which measures how often predicted LTV falls into the same value bucket as the true LTV ranking, and (ii) \textbf{MAE} within each bucket to quantify prediction error.

\begin{figure}[!htbp]
  \centering
  \begin{subfigure}[t]{0.49\linewidth}
    \centering
    \includegraphics[width=\linewidth]{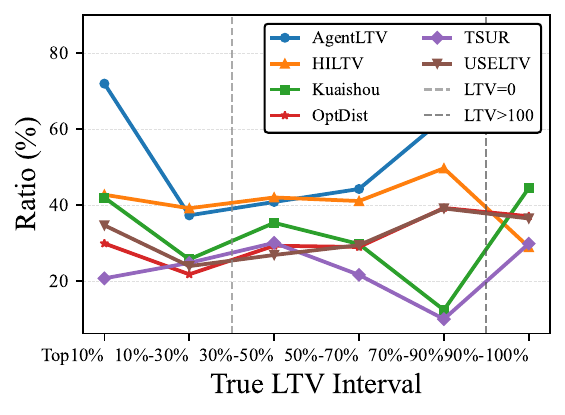}
    \caption{Same Interval Ratio.}
    \Description{Same Interval Ratio.}
    \label{fig:Same Interval Ratio}
  \end{subfigure}
  \begin{subfigure}[t]{0.49\linewidth}
    \centering
    \includegraphics[width=\linewidth]{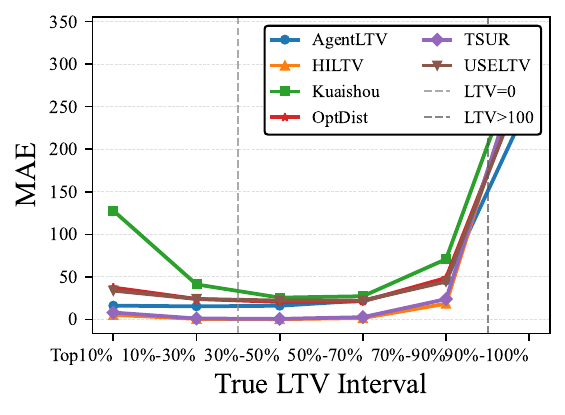}
    \caption{Mean Absolute Error.}
    \Description{Mean Absolute Error.}
    \label{fig:Economic Analysis}
  \end{subfigure}

  \caption{Online economic analysis on bucketed LTV values. 
  (a) Same Interval Ratio: the fraction of users whose predicted LTV falls into the same value bucket as the bucket defined by the ground-truth LTV ranking. 
  Higher is better and indicates stronger ranking consistency across buckets. 
  (b) Mean Absolute Error (MAE) within each bucket: the absolute prediction error averaged over users in that bucket. 
  Lower is better and reflects better value calibration.}
  \label{fig:online_economic_analysis}
\end{figure}

Figure~\ref{fig:Same Interval Ratio} shows that AgentLTV achieves strong ranking quality across buckets. 
The advantage is most clear for the high-value bucket and the negative-LTV bucket. 
This suggests that the discovered pipeline can better separate high-potential users from high-risk users under realistic traffic. 
Figure~\ref{fig:Economic Analysis} further shows lower MAE in most buckets, indicating improved calibration of predicted values.


\subsection{Public Dataset Experiment}

To answer RQ4, we evaluate AgentLTV on the \textit{Kaggle Acquire Valued Shoppers Challenge} dataset\footnote{https://www.kaggle.com/competitions/acquire-valued-shoppers-challenge}, following the same protocol as ZILN~\cite{wang2019deep}. 
The task is to predict each user's total purchase value within 12 months after the first transaction.

\begin{table}[H]
  \caption{Basic statistics of the public dataset.}
  \label{tab:open data statistics}{\footnotesize
  \begin{tabular}{cccc}
    \toprule
     Samples& Standard deviation& Average LTV\\
    \midrule
     1,797,813 & 6,452.395& 797.358\\
  \bottomrule
\end{tabular}}
\end{table}

We use initial purchase statistics and categorical attributes (e.g., chain, category, brand, and product size). 
We focus on the top 20 companies by customer volume and select customers whose first purchase falls between March 1, 2012, and July 1, 2012. 
We split the data into train/validation/test with an 8:1:1 ratio. 
Table~\ref{tab:open data statistics} summarizes basic statistics.

\begin{table}[!htbp]
  \caption{Results of comparative experiments on public datasets.}
  \label{tab:opendata_cpmparative}
  {\footnotesize
  \begin{tabular}{ccccc}
    \toprule
        Method       & ER   & Norm GINI & Spearman & RMSE  \\  
      \midrule
      Wide\&Deep   & 0.522  & 0.695       & 0.611      & 6306.097   \\
      DeepFM       & 0.533  & 0.697       & 0.615      & 4566.259   \\
      ZILN         & 0.504  & 0.726       & 0.645      & 4879.940   \\
      DCN          & 0.619  & 0.507       & 0.613      & 10833.894   \\
      GateNet      & 0.480  & 0.725       & 0.640      & 4938.126   \\
      Kuaishou     & 0.542  & 0.728       & 0.645      & 9138.638   \\
      TSUR         & 0.479  & 0.711       & 0.632      & 5237.200   \\
      OptDist      & 0.510  & 0.720       & 0.638      & 6416.406   \\
      USE-LTV      & 0.503  & 0.725       & 0.643      & 6397.289   \\
      Hi-LTV       & 0.898  & 0.712       & 0.637      & 6457.647   \\
      AgentLTV  & \textbf{0.479}  & \textbf{0.732} & \textbf{0.649}   & \textbf{3646.851}   \\
  \bottomrule
\end{tabular}}
\end{table}

Table~\ref{tab:opendata_cpmparative} reports the results. 
AgentLTV consistently outperforms the strongest baselines across all metrics. 
We open-source our processed data and the discovered program at GitHub\footnote{https://github.com/wuchaowei123/AgentLTV}. 
Interestingly, the discovered model shares key design choices with ZILN~\cite{wang2019deep}, suggesting that AgentLTV can rediscover effective LTV modeling patterns on public benchmarks.


\section{Conclusion}

This study presents AgentLTV, a general multi-agent-driven framework for automated LTV modeling. It targets practical challenges in the labor-intensive LTV development cycle, including data cleaning, feature engineering, objective design, and hyperparameter tuning, by automating a search over executable pipeline programs, thereby improving reuse across scenarios. AgentLTV combines MCTS for broad exploration with EA for refinement. Such decision agents coordinate the search, while auxiliary agents support auto-coding, execution-based evaluation, and code repair. The resulting framework reduces manual trial-and-error, adapts to shifting data patterns and business constraints, and supports LTV settings with negative values common in cost-dominated acquisition.


Experiments on a large-scale industrial dataset and a public benchmark show that AgentLTV can discover competitive models with strong ranking and error performance. 
The results also validate the roles of the MCTS and EA modules in exploration and refinement.

In future work, we will extend AgentLTV with a dedicated agent for data profiling and preprocessing. 
This agent will automatically summarize dataset characteristics and feed structured signals into the search space. 
We also plan to improve the safety and stability of auto-generated code via stronger interface checks and more reliable execution traces. 
Although this work focuses on LTV, the same agentic search-and-evolve paradigm may apply to other structured prediction pipelines that require heavy engineering.

\bibliographystyle{ACM-Reference-Format}
\bibliography{sample-base}

\appendix

\section{Agent Prompts}\label{app:Agent Prompts}
This appendix summarizes the core prompt designs used by different agent in our system. 
\subsection{Code Generator}
\begin{lstlisting}
f"""You are implementing an algorithm for: {task_name}
## Your Task:
{method_hint if method_hint else "Improve this code to achieve a better score."}
## CRITICAL REQUIREMENTS:
1. Replace the main algorithm function with the specified method
2. Do NOT modify scoring/evaluation functions
3. Preserve output format: `print(f"score = {{score}}")`
4. The code must be complete and executable
## You MUST:
- Understand the provided template and problem structure
- Implement the algorithm described in "Your Task"
- Replace all placeholder/baseline logic
- Validate collisions or constraints if applicable (use template helpers)
## You MUST NOT:
- Change scoring functions
- Change output format
- Modify problem class definitions (e.g., data structures)
## Output:
Generate complete, self-contained Python code implementing your algorithm."""
\end{lstlisting}

\subsection{Advice Generator}
\begin{lstlisting}
COMPARISON_PROMPT = """Compare these two code solutions to the same problem.
Explain the main principles that differ between the codes:
CODE 1:```python{code1}```
CODE 2:```python{code2}```
Please identify:
1. The main algorithmic approaches used in each
2. Key differences in methodology
3. Strengths of each approach"""

HYBRID_PROMPT = """We have up until now done experiments with two major types of codes, that are described in detail below.
### Comparison Analysis
{comparison}
### CODE 1 (Score: {score1})```python{code1}```
### CODE 2 (Score: {score2})```python{code2}```
PLEASE CREATE AN ALGORITHM THAT USES THE BEST PARTS OF BOTH STRATEGIES 
TO CREATE A HYBRID STRATEGY THAT IS TRULY WONDERFUL AND SCORES HIGHER 
THAN EITHER OF THE INDIVIDUAL STRATEGIES.
Please provide the hybrid code:
```python# YOUR HYBRID CODE```"""

FORMAT_IDEA_PROMPT = """Structure the given idea into the following format:
<description>
Your description about the method goes here.
</description>
<steps>
Your list of steps to implement the method goes here.
</steps>
<notes>
Strengths and weaknesses of the idea goes here.
</notes>
Idea to format:
{idea}"""
"""
\end{lstlisting}

\subsection{Expert Optimizer}
\begin{lstlisting}
prompt = f"""Analyze the following Python code and select the most appropriate expert advice type.
### Code to Analyze:
```python
```
### Available Advice Types:
{advice_options}
### Task:
Based on the code's purpose, libraries used, and domain, select the SINGLE most appropriate advice type.
Respond with ONLY the number (1-{len(self.advice_library)}) of the best advice type, nothing else.
For example, if boosted_trees is best, respond with just: 2"""

prompt = f"""### Expert Advice ({advice.name})
{advice.advice}
### Current Code
```python
{node.code}
```
### Task
Apply the expert advice above to improve the code. Focus on:
1. Following the advice guidelines
2. Maintaining correctness
3. Improving the score/performance
Please provide the improved code:
```python # YOUR IMPROVED CODE```"""
\end{lstlisting}

\section{LTV Domain Knowledge}\label{app:LTV Domain knowledge}
In this section, we report the domain knowledge provided to LLMs for generating initial nodes of Monte Carlo trees about LTV modeling. Due to space limitations, we will only provide a brief description of method 1, and the other methods are omitted.
\begin{lstlisting}
**Method 1: Custom AWMSE LightGBM (Gradient/Hessian Derivation)**
<description>
This method involves the analytical derivation of the first derivative (Gradient $g$) and second derivative (Hessian $h$) of the Asymmetric Weighted MSE (AWMSE) objective function to enable intrinsic optimization within Gradient Boosting Decision Tree (GBDT) frameworks. The objective function $\mathcal{L}(\hat{y})$ is piecewise defined based on the relationship between the prediction $\hat{y}$ and the true value $y$. By providing these custom derivatives, the model can inherently prioritize the minimization of high-cost errors during training rather than relying on post-hoc adjustments.
**Method 2: Multi-Task DNN with ZILN loss**
**Method 3: Risk-Averse Quantile Regression**
**Method 4: Three-Part Asymmetric Hurdle Framework**
**Method 5: Temporal Convolutional Networks (TCN)Encoder**
**Method 6: FT-Transformer for Feature-Wise Attention**
**Method 7: High-Cardinality Categorical Feature Strategy (Dual Encoding)**
**Method 8: Hybrid Feature Fusion GBDT**
**Method 9: Level 2 Stacking with AWMSE Meta ptimization**
**Method 10: Dynamic Conservative Prediction Calibration**
**Method 11: Multi-Task Learning (MTL) DNN with Sign-Regularization**
**Method 12: Feature Engineering- 7-Day Temporal Compression Toolkit**
**Method 13: Bidirectional LSTM with Attention for Temporal Risk Modeling**
**Method 14: TabNet for Self-Supervised Feature Selection**
**Method 15: CatBoost with Ordered Boosting and Custom Asymmetric Loss**
**Method 16: Graph Neural Network (GNN) for User Similarity Modeling**
**Method 18: Temporal Fusion Transformer (TFT) for Multi-Horizon LTV Forecasting**
**Method 19: Ridge/Linear Regression Baseline**
**Method 20: XGBoost with Custom Asymmetric Loss**
**Method 21: Logistic Regression + Hurdle Framework**
**Method 22: Random Forest with Feature Importance Analysis**
**Method 23: MLP-Mixer for Tabular Data**
**Method 24: Retrieval-Augmented LTV (RALTV)**
**Method 25: Diffusion Model for LTV Distribution Estimation**
**Method 26: Mixture of Experts (MoE) for User Segmentation**
**Method 27: Reinforcement Learning for LTV**
**other 7 methods**
\end{lstlisting}

\section{EA Module Search Process}\label{app:search result}
In this section, we report the top 20 LTV prediction models generated by the evolutionary algorithm module based on the optimal code of the MCTS module, as shown in Table~\ref{tab:bt_ios_pltv_top20}.

\begin{table*}[t]
\centering
\caption{EA Module top 20 LTV Model Results Ranked by Node Value}
\label{tab:bt_ios_pltv_top20}
\scriptsize
\resizebox{\textwidth}{!}{%
\begin{tabular}{@{}rllccccll@{}}
\toprule
\textbf{Rank} & \textbf{Gen} & \textbf{Node Value} & \textbf{Gini} & \textbf{Spearman} & \textbf{RMSE} & \textbf{ErrorRate} & \textbf{Evolution type} & \textbf{Method} \\
\midrule
1 & gen\_21 & 0.834602 & 0.9557 & 0.8140 & 97.71 & 0.7726 & Crossover & enhanced\_temporal\_attention\_mechanism \\
2 & gen\_4 & 0.832967 & 0.9540 & 0.8272 & 104.13 & 0.9014 & Crossover & enhance\_tcn\_architecture\_with\_attention \\
3 & gen\_19 & 0.821321 & 0.9532 & 0.8168 & 111.36 & 0.6851 & Crossover & hybrid\_dual\_branch\_attention\_tcn \\
4 & gen\_18 & 0.819654 & 0.9508 & 0.8292 & 117.27 & 0.7690 & Crossover & kinematic\_attention\_mtl\_rank\_boost \\
5 & gen\_45 & 0.788169 & 0.9586 & 0.8148 & 125.33 & 0.6690 & Mutation & log\_cosh\_robust\_mtl\_with\_enhanced\_features \\
6 & gen\_35 & 0.786940 & 0.9519 & 0.8148 & 124.57 & 0.9622 & Crossover & ranking\_loss\_and\_attention \\
7 & gen\_0 & 0.784552 & 0.9578 & 0.7980 & 122.01 & 0.5243 & Init & initial\_program \\
8 & gen\_43 & 0.783808 & 0.9512 & 0.8131 & 126.80 & 1.0304 & Mutation & mutual\_dml\_tcn \\
9 & gen\_2 & 0.776483 & 0.9485 & 0.8440 & 145.69 & 0.6710 & Mutation & attentive\_rank\_mtl \\
10 & gen\_25 & 0.774645 & 0.9627 & 0.8173 & 141.09 & 0.5947 & Crossover & add\_static\_volatility\_trend \\
11 & gen\_14 & 0.763827 & 0.9644 & 0.8179 & 152.65 & 0.8469 & Crossover & add\_pearson\_loss \\
12 & gen\_3 & 0.758138 & 0.9623 & 0.8086 & 153.97 & 0.6200 & Crossover & enhanced\_tcn\_depth\_and\_attention \\
13 & gen\_9 & 0.753085 & 0.9631 & 0.8237 & 165.32 & 0.6561 & Mutation & kinematic\_hybrid\_mtl \\
14 & gen\_6 & 0.748945 & 0.9584 & 0.8042 & 160.26 & 0.8756 & Crossover & arch\_se\_acc\_std \\
15 & gen\_22 & 0.748545 & 0.9618 & 0.8302 & 172.08 & 0.6365 & Crossover & se\_att\_accel\_arch \\
16 & gen\_15 & 0.736407 & 0.9442 & 0.8025 & 167.83 & 0.7067 & Crossover & kinematic\_dual\_branch\_network\_with\_pearson\_loss \\
17 & gen\_23 & 0.734145 & 0.9633 & 0.8682 & 222.12 & 0.7198 & Crossover & tail\_huber\_loss \\
18 & gen\_16 & 0.721330 & 0.9547 & 0.8152 & 191.16 & 0.7903 & Crossover & attn\_pool\_fusion \\
19 & gen\_11 & 0.638605 & 0.9565 & 0.8340 & 281.93 & 0.9373 & Crossover & add\_softclip\_helper \\
20 & gen\_20 & 0.608377 & 0.9066 & 0.6541 & 225.25 & 0.9730 & Mutation & transinf\_heteroscedastic\_mtl \\
\bottomrule
\end{tabular}%
}
\end{table*}

\section{Our Optimal LTV Prediction Model}\label{app:our best model}
This section describes the optimal LTV prediction model in Algorithm~\ref{alg:AgentLTV}. The final model is a multi-branch network with a multi-task output tower. 
For dynamic features, we first project them into a high-dimensional space and feed them into a temporal convolution branch and an attention branch. 
For static features, we use embedding layers followed by an MLP branch. 
We then pool and fuse the three branch representations and apply an MLP head to produce multi-task predictions, including LTV. Note that the discovered model remains lightweight, which facilitates online deployment and maintenance.

\begin{algorithm}[H]
\caption{The optimal model generated by AgentLTV}
\label{alg:AgentLTV}
\fontsize{8pt}{10pt}\selectfont 
\begin{algorithmic}[1]
\REQUIRE 
  $D_{\text{tr/va/te}}$: Train/val/test datasets; $y$: Target (REC\_USD\_D60); \\
  \quad $\mathcal{H}$: Hyperparams ($d_{\text{model}}=192$, $\text{depth}_{\text{TCN}}=4$, $\text{depth}_{\text{Attn}}=2$, batch=512)
\ENSURE 
  $\hat{y}_{\text{te}}$: Test predictions; \\
  \quad $S$: Pareto-optimized score (Gini/Spearman/RMSE/error rate)

\STATE \textbf{1. Kinematic Feature Engineering}
\STATE $X_{\text{raw}} \in \mathbb{R}^{|\mathcal{U}| \times 7 \times D}$: 7-day numeric tensors; $\mathcal{M}$: validity mask
\STATE $\tilde{X}_{\text{seq}} \leftarrow$ Masked standardization of $X_{\text{raw}}$
\STATE $X_{\text{seq}} \leftarrow \text{Concat}(\tilde{X}_{\text{raw}}, \Delta\tilde{X}, \Delta^2\tilde{X}, \sum \tilde{X}_{\text{raw}}\mathcal{M})$ 
  \COMMENT{$\Delta$: velocity; $\Delta^2$: acceleration; $\sum$: integral}
\STATE $X_{\text{static/cat}}$, $\mathcal{E}$: Static/categorical features + embeddings

\STATE \textbf{2. Dual-Branch Network Training}
\STATE Init $\text{KinematicHybridNet}$:
  \begin{itemize}
  \item TCN Branch: 4 residual blocks (kernel=3, dilation=$2^i$, GELU)
  \item Attn Branch: 2 multi-head blocks (nhead=4, mask=$\neg\mathcal{M}$)
  \item Fusion: Concat(TCN/Attn mean, static emb) $\to$ MLP($d_{\text{model}}$)
  \item Heads: $\hat{y}_{\text{mean}}$, $\hat{y}_{\text{q20}}$, $\text{logit}(P(y<0))$
  \end{itemize}
\STATE Loss ($\lambda=20$):
  \[
  \mathcal{L} = \text{AWMSE} + 0.35(\text{PinballLoss} + \text{BCE}) + \lambda(1-\rho)
  \]
  \COMMENT{$\rho$: Pearson corr; BCE: pos-weighted}
\STATE Optimize (AdamW, lr=8e-4, 45 epochs, early stop)

\STATE \textbf{3. Policy Tuning \& Inference}
\STATE Predict $\hat{y}_{\text{mean/q20}}^{\text{va}}$, $P_{\text{neg}}^{\text{va}}$ on $D_{\text{va}}$
\STATE $\theta^* = \arg\max_{\theta} S(\text{ApplyPolicy}(\hat{y}^{\text{va}},\theta),y_{\text{va}})$
\STATE Calibrate: $\hat{y}_{\text{cal}}^{\text{va}} = a\hat{y}_{\text{raw}}^{\text{va}}+b + (\mu_{\text{anchor}} - \mathbb{E}[\hat{y}_{\text{raw}}^{\text{va}}])$
\STATE $\hat{y}_{\text{te}} = (a\cdot\text{ApplyPolicy}(\hat{y}^{\text{te}},\theta^*)+b)-\delta^*$; $S = \text{ComputeScore}(y_{\text{te}},\hat{y}_{\text{te}})$
\RETURN $S, \hat{y}_{\text{te}}$
\end{algorithmic}
\fontsize{10pt}{12pt}\selectfont 
\end{algorithm}

\clearpage  
\balance    
\end{document}